\documentclass[10pt,twocolumn,letterpaper]{article}  

\usepackage{amsmath,amsfonts,bm}

\def\eqref#1{equation~\ref{#1}}

\def\1{\bm{1}}

\DeclareMathAlphabet{\mathsfit}{\encodingdefault}{\sfdefault}{m}{sl}
\SetMathAlphabet{\mathsfit}{bold}{\encodingdefault}{\sfdefault}{bx}{n}

\usepackage{iccv}
\usepackage{times}
\usepackage{epsfig}
\usepackage{graphicx}
\usepackage{amsmath}
\usepackage{amssymb}
\usepackage{enumitem}
\usepackage{url}
\usepackage{color,xcolor}
\usepackage{textcomp,gensymb}
\usepackage{booktabs}
\usepackage{multirow}
\usepackage{tabu}
\usepackage[super]{nth}
\usepackage{subfig}
\usepackage{adjustbox}
\usepackage{wasysym}
\usepackage{pifont}
\usepackage{caption}
\usepackage{xspace}

\usepackage[colorlinks = true,
            linkcolor = purple,
            urlcolor  = magenta,
            citecolor = green,
            anchorcolor = blue]{hyperref}

\definecolor{myblue}{HTML}{3E3EE9}
\definecolor{myorange}{HTML}{FF9000}
\definecolor{mygreen}{HTML}{39BE28}
\definecolor{myyellow}{HTML}{D1CA00}
\newcommand{\cmark}{\ding{51}}
\newcommand{\xmark}{\ding{55}}

\newcommand{\ourwork}{DualCross\xspace}

\PassOptionsToPackage{table}{xcolor}
\iccvfinalcopy 
\begin{document}

\title{\ourwork: Cross-Modality Cross-Domain Adaptation for Monocular BEV Perception}

\author{Yunze Man, Liang-Yan Gui, Yu-Xiong Wang\\
{University of Illinois at Urbana-Champaign Champaign, IL 61820, USA} \\ 
{\tt\small \{yunzem2, lgui, yxw\}@illinois.edu}}

\author{Yunze Man\\
UIUC\\
{\tt\small yunzem2@illinois.edu}
\and
Liang-Yan Gui\\
UIUC\\
{\tt\small lgui@illinois.edu}
\and
Yu-Xiong Wang\\
UIUC\\
{\tt\small yxw@illinois.edu}
}

\twocolumn[{%
\renewcommand\twocolumn[1][]{#1}%
\maketitle
\begin{center}
    \centering
    \captionsetup{type=figure}
    \includegraphics[trim=39 162 33 147, clip=True, width=1\textwidth]{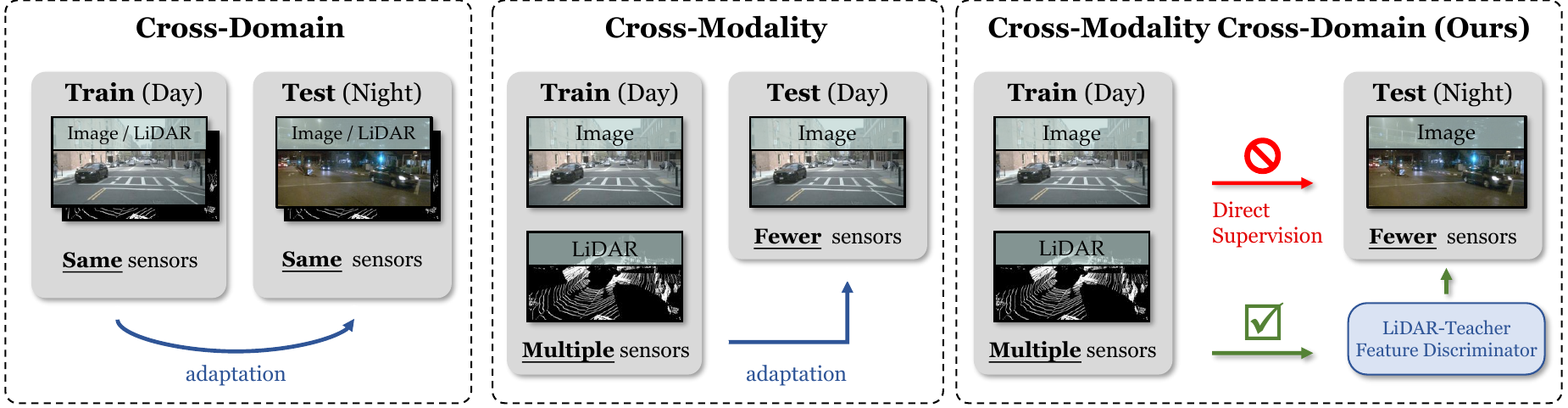}
    \vspace{-4mm}
    \captionof{figure}{\textit{Left \& Middle}: Existing adaptation models assume either a fixed modality or a fixed domain between training and testing phases. \textit{Right}: A more realistic setting considers both cross-modality and cross-domain shifts. We propose \ourwork to reduce the domain and modality discrepancy, and achieve state-of-the-art performance. 
    }
    \label{fig:teaser}
\end{center}
}]
\ificcvfinal\thispagestyle{empty}\fi

\begin{abstract}

Closing the domain gap between training and deployment and incorporating multiple sensor modalities are two challenging yet critical topics for self-driving. Existing work only focuses on single one of the above topics, overlooking the simultaneous domain and modality shift which pervasively exists in real-world scenarios. A model trained with multi-sensor data collected in Europe may need to run in Asia with a subset of input sensors available.  In this work, we propose \ourwork, a cross-modality cross-domain adaptation framework to facilitate the learning of a more robust monocular bird's-eye-view (BEV) perception model, which transfers the point cloud knowledge from a LiDAR sensor in one domain during the training phase to the camera-only testing scenario in a different domain. This work results in the first open analysis of cross-domain cross-sensor perception and adaptation for monocular 3D tasks in the wild. We benchmark our approach on large-scale datasets under a wide range of domain shifts and show state-of-the-art results against various baselines. Our project webpage is at \href{https://yunzeman.github.io/DualCross}{https://yunzeman.github.io/DualCross}.

\end{abstract}

\section{Introduction} \label{sec:intro}

In recent years, multi-modality 3D perception has shown outstanding performance and robustness over its single-modality counterpart, achieving leading results for various 3D perception tasks~\cite{jaritz2020xmuda,man2021multi,park2021multi,qcl20imvotenet,vlh20pointpainting,weng2020gnn3dmot} on large-scale multi-sensor 3D datasets~\cite{Caesar2019nuScenes,lyftdataset,sun2020Waymo}. Despite the superiority in information coverage, the introduction of more sensor modalities also poses additional challenges to the perception system. On one hand, generalizing the model between datasets becomes hard, because each sensor has its unique properties, such as field-of-view (FoV) for cameras, density for LiDAR, \etc. On the other hand, the operation of the model is conditioned on the presence and function of more sensors, making it hard to work on autonomous agents with less sensor types or under sensor failure scenarios.

More specifically, transferring knowledge among different data domains is still an open problem for autonomous agents in the wild. In the self-driving scenario, training the perception models offline in a source domain with annotation while deploying the model in a different target domain without annotation is very common in practice. As a result, a model has to consider the domain gap between source and target environments or datasets, which usually involves different running locations, different sensor specifications, different illumination and weather conditions, \etc. 

Meanwhile, in addition to domain shift, modality shift is another factor which challenges the successful deployment of models. The widely adopted assumption that all sensors are available during training, validation, and deployment time is not always true in reality. Due to the cost and efficiency trade-off, or sensor missing and failure, in many scenarios we can have fewer sensors available in the target domain during testing than what we have in the source domain during training. A typical scenario is having camera and LiDAR sensors in the large-scale training phase while only having cameras for testing, as shown in Figure~\ref{fig:teaser}. It is not clear how to facilitate the camera-only 3D inference with the help of a LiDAR sensor only in the source domain during training.

The challenges above raise an important question: {\it Can we achieve robust 3D perception under both domain shift and sensor modality shift?} Existing methods either study cross-domain scenarios assuming consistent modality~\cite{gong2021mdalu,jaritz2020xmuda,kundu2018adadepth,li2021cross,luo2021unsupervised,peng2021sparse,qin2019pointdan,yang2021st3d,zhang2021srdan}, or study cross-modality scenarios assuming the same domain during training and validation~\cite{BEVDistill,MonoDistill,Liga-stereo,CMKD,li2022unifying,li2022bevdepth,yan20222dpass}. However, simultaneous domain and modality shift poses additional challenges of large domain discrepancy and exacerbates the ill-pose nature of 3D inference from monocular information due to the misaligned sensory data. As we will discuss in Sec.~\ref{sec:method_teacher}, our new problem setting requires a novel methodology in using LiDAR without increasing the domain discrepancy. 

To tackle the above challenges, we propose \ourwork, a cross-modality cross-domain adaptation framework for bird's-eye-view (BEV) perception. Our model addresses the monocular 3D perception task between different domains, and utilizes additional modalities in the source domain to facilitate the evaluation performance. Motivated by the fact that image and BEV frames are bridged with 3D representation, we first design an efficient backbone to perform 3D depth estimation followed by a BEV projection. Then, to learn from point clouds without explicitly taking them as model inputs, we propose an implicit learning strategy, which distills 3D knowledge from a LiDAR-Teacher to help the Camera-Student learn better 3D representation. Finally, in order to address the visual domain shift, we introduce adversarial learning on the student to align the features learned from source and target domains. Supervision from the teacher and feature discriminators are designed at multiple layers to ensure an effective knowledge transfer. 

By considering the domain gap and effectively leveraging LiDAR point clouds in the source domain, our proposed method is able to work reliably in more complicated, uncommon, and even unseen environments. Our model achieves state-of-the-art performance in four very different domain shift settings. Extensive ablation studies are conducted to investigate the contribution of our proposed components, the robustness under different changes, and other design choices.

The main contributions of this paper are as follows. (1) We introduce mixed domain and modality mismatch, an overlooked but realistic problem setting in 3D domain adaptation in the wild, leading to a robust camera-only 3D model that works in complicated and dynamic scenarios with minimum sensors available. (2) We propose a novel LiDAR-Teacher and Camera-Student knowledge distillation model, which considerably outperforms state-of-the-art LiDAR supervision methods.  (3) Extensive experiments in challenging domain shift settings demonstrate the capability of our methods in leveraging source domain point cloud information for accurate monocular 3D perception.
\section{Related Work} \label{sec:related}

\noindent\textbf{Multi- and Cross-modality 3D Perception.} Considerable research has examined leveraging signals from multiple modalities, especially images and point clouds, for 3D perception tasks. Early work~\cite{lyc19mtms} projects point clouds to the BEV frame and fuses 2D RGB features to generate proposals and regress bounding boxes. Later work~\cite{ykk203dcvf,zmp20cross} explores deep fusion between points and images. Under the umbrella of the cross-modality setting, 2DPASS~\cite{yan20222dpass} transfers features learned from images to the LiDAR. BEVDepth~\cite{li2022bevdepth} obtains reliable depth estimation by exploiting camera parameters with image features during training. More recently, a line of work explores knowledge distillation from one sensor to another for 3D object detection~\cite{BEVDistill,MonoDistill,Liga-stereo,CMKD,li2022unifying,li2022bevdepth,yan20222dpass}. On the contrary, our method explores a more realistic yet challenging setting, where we use LiDAR data in one domain (Boston/Sunny/Daylight) during training to help the camera-only model during inference in another domain (Singapore/Rainy/Night). As a result, we analyze and improve the actual usefulness of additional sensors under domain shift settings.

\vspace{2mm}\noindent\textbf{Cross-domain 3D Perception.} While extensive research has been conducted on domain adaptation for 2D tasks, the field of domain adaptation for 3D perception in the real world has received relatively less attention. Some prior work adapts depth estimation from synthetic to real image domains~\cite{kundu2018adadepth,zhao2019geometry}. Working on point clouds, PointDAN~\cite{qin2019pointdan} designs a multi-scale adaptation model for 3D classification. For 3D semantic segmentation, SqueezeSeg~\cite{wu2019squeezesegv2} projects point clouds to the 2D view, while other work~\cite{gong2021mdalu,jaritz2020xmuda,peng2021sparse} leverages point clouds and images data together. Recent work~\cite{luo2021unsupervised,yang2021st3d,zhang2021srdan} explores cross-domain 3D object detection from point clouds. SRDAN~\cite{zhang2021srdan} employs adversarial learning to align the features between different domains. Although prior work~\cite{jaritz2020xmuda,li2021cross} explores various domain adaptation techniques for different sensor modalities, these methods only adopt the same modalities to learn the domain shift between source and target data. In contrast, our approach achieves robust 3D perception in a more general scenario, where the model can perform accurate 3D inference in the target domain by adapting information encoded in source-exclusive modalities. 

\vspace{2mm}\noindent\textbf{3D Inference in Bird's-Eye-View Frame.} Inferring 3D scenes from the BEV perspective has recently received a large amount of interest due to its effectiveness. MonoLayout~\cite{mani2020monolayout} estimates the layout of urban driving scenes from images in the BEV frame and uses an adversarial loss to enhance the learning of hidden objects. Another work~\cite{can2021structured} proposes to employ graphical representation and temporal aggregation for better inference in the driving scenarios using on-board cameras. Recently, using BEV representation to merge images from multiple camera sensors has become a popular approach~\cite{hendy2020fishing,pan2020cross}. Following the monocular feature projection proposed by Orthographic Feature Transform (OFT)~\cite{roddick2018orthographic}, Lift-Splat-Shoot~\cite{philion2020lift} disentangles feature learning and depth inference by learning a depth distribution over pixels to convert camera image features into BEV. 
Unlike the above work performing BEV analysis in settings with more controlled premises, we are the first to explore cross-domain and cross-sensor settings, leading to a more robust and more realistic 3D inference methodology.
\begin{figure*}[!t]
    \centering
    \vspace{2.5mm}
    \includegraphics[trim=0 60 0 115,clip, width=0.98\linewidth]{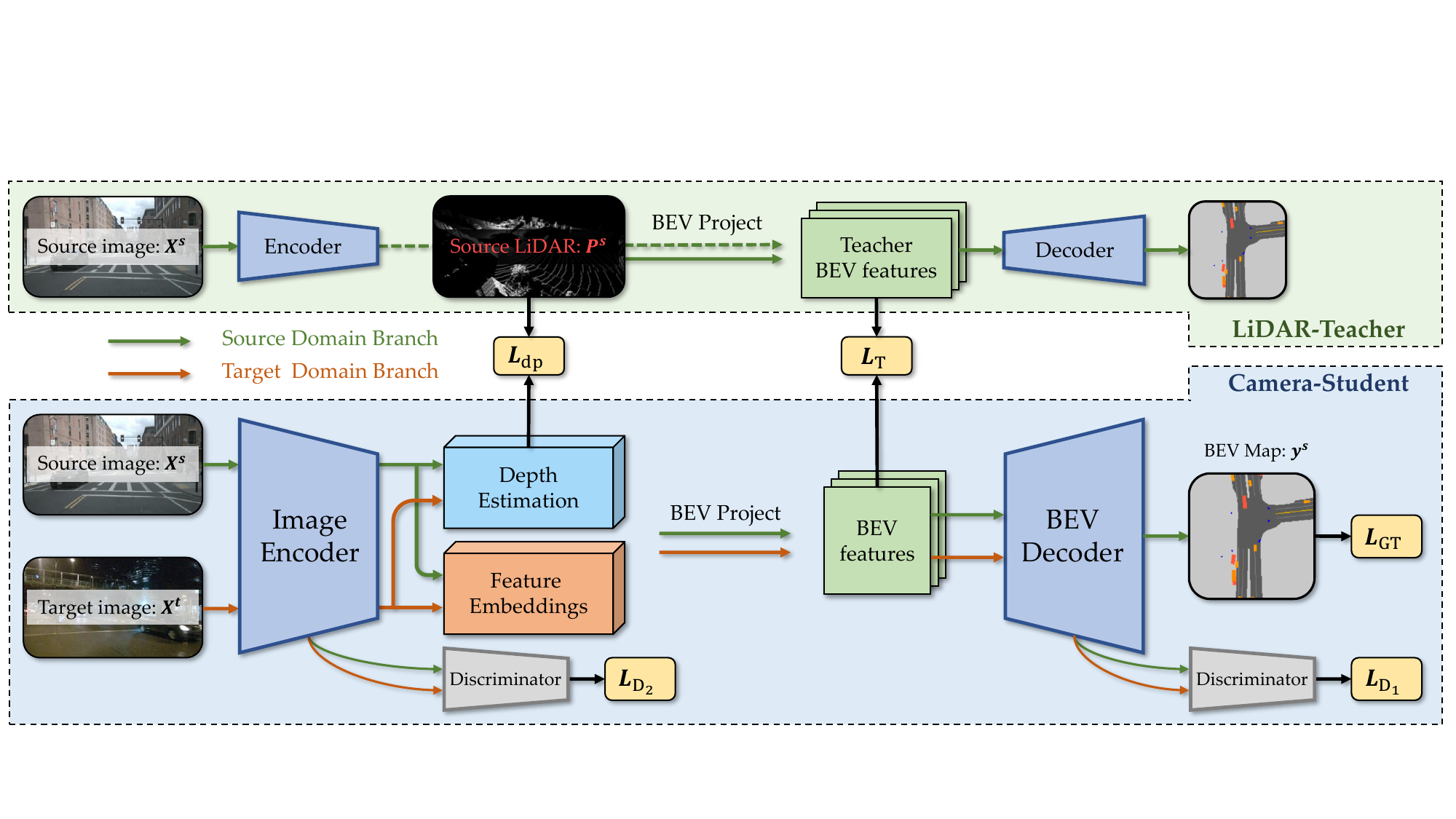}
    \vspace{-0.5mm}
\caption{\textbf{Overview of our \ourwork framework}. \ourwork includes three components. (1) \textbf{LiDAR-Teacher} uses voxelized LiDAR point clouds to transform the image features to the BEV frame. It provides essential knowledge on how to guide image learning given LiDAR information. (2) \textbf{Camera-Student} is supervised by the teacher model as well as the LiDAR ground truth. (3) \textbf{Discriminators} are used to align features from source and target domains.}
    \label{fig:main}
    \vspace{-5mm}
\end{figure*}

\section{Approach} \label{sec:method}

In this work, we consider the task of learning BEV representation of scenes with domain shift and modality mismatch. Specifically, the model will be given annotated LiDAR point clouds and camera images in the source domain, but only unannotated camera images in the target domain. And the model seeks to achieve highest performance on the unsupervised target domain. This problem setting is common and worthwhile, especially considering the existence of many existing public multi-modality datasets and the rise of many camera-only vehicle scenarios.

Formally, \textbf{for the source domain}, we are given \textit{labeled} data with $N^{s}$ multi-modality samples, $\mathcal{D}^s=\{(\boldsymbol{X}^{s}_{i},\boldsymbol{P}^{s}_{i},\boldsymbol{y}^{s}_{i})\}_{i=1}^{N^{s}}$, where $s$ represents the source domain. Here $\boldsymbol{X}^{s}_{i}=\{\boldsymbol{x}^{s}_{ik}\}_{k=1}^{n}$ consists of $n$ camera images $\boldsymbol{x}^{s}_{ik} \in \mathbb{R}^{3\times H\times W}$. The number of cameras $n$ can take any integer as small as one, depending on the dataset or cameras deployed on the vehicle. In addition, each camera image has an intrinsic matrix and an extrinsic matrix. $\boldsymbol{P}^{s}_{i}$ is a point cloud containing multiple unordered points $\boldsymbol{p}\in \mathbb{R}^3$ represented by 3D coordinate values. And label $\boldsymbol{y}^s_i$ represents rasterized representation of the scenes in the BEV coordinate. \textbf{For the target domain}, we are given \textit{unlabeled} data with $N^{t}$ image samples, $\mathcal{D}^t=\{\boldsymbol{X}^{t}_{i}\}_{i=1}^{N^{t}}$, where $t$ represents the target domain, and we want to estimate $\{\boldsymbol{y}^t_i\}_{i=1}^{N^{t}}$, the BEV representation of the scenes in the target domain.

An overview of our method \ourwork is illustrated in Figure~\ref{fig:main}. \ourwork is designed to extract features from monocular images and project the features into the BEV frame (Section~\ref{sec:method_bev}), using estimated or ground truth 3D depth information. The model is composed of a LiDAR-teacher and a Camera-student (Section~\ref{sec:method_teacher}), where the teacher encodes how to learn better representation given point clouds, and transfers that knowledge to the camera-only student using multi-level teacher-student supervision. Finally, to bridge the domain gap between source and target domains, we leverage adversarial discriminators at different feature layers to align the distributions across two domains in the camera-student model (Section~\ref{sec:method_adv}). Finally, we describe the overall learning objective and loss designs (Section~\ref{sec:method_loss}).

\subsection{Learning BEV from Images} \label{sec:method_bev}

In order to achieve 3D perception under the cross-modality setting, our first challenge is to unify the image coordinates, point cloud coordinates, and BEV coordinates into a joint space. We follow LSS~\cite{philion2020lift} to transform the image features from perspective view into the BEV view. Specifically, we tackle this problem by constructing a 3D voxel representation of the scene for each input image. We discretize the depth axis into $N_d$ bins and lift each pixel of the images into multiple voxels (frustums), where each voxel is represented by the 3D coordinate of its center location. For a given pixel $\mathrm{px}=(h, w)$ on one of the camera image, it corresponds to a set of $N_d$ voxels at different depth bins:

{\small
  \setlength{\abovedisplayskip}{-4pt}
  \setlength{\belowdisplayskip}{6pt}
  \setlength{\abovedisplayshortskip}{0pt}
  \setlength{\belowdisplayshortskip}{3pt}
\begin{align}\label{eq:feature_voxel}
    V_{\mathrm{px}}=\{v_i = M^{-1}[d_ih, d_iw, d_i]^T | i \in \{1, 2, \cdots, N_d\}\},
\end{align}}%
where $M$ is camera matrix and $d_i$ is the depth of the $i$-th depth bin. The feature vector of each voxel $v_i$ in $V_{\mathrm{px}}$ is the base feature $\boldsymbol{f}_{\mathrm{px}}$ of pixel $\mathrm{px}$ scaled by the depth value $\alpha_{i}$. More specifically, $\boldsymbol{f}_{v_i \in V_{\mathrm{px}}} = \alpha_i \cdot \boldsymbol{f}_{\mathrm{px}}$, where the pixel feature $\boldsymbol{f}_{\mathrm{px}}$ is extracted by an image encoder. And the depth value $\alpha_{i}$ is obtained either from LiDAR point clouds or by estimation, in the teacher and student model, respectively. The acquirement of $\alpha_{i}$ is introduced in Sec.~\ref{sec:method_teacher}. 

After getting the feature for each of the voxels, we project the voxels onto the BEV and aggregate the features to get the BEV feature map. The BEV frame is rasterized into $(X, Y)$ 2D grids, and for each grid, its feature is constructed from the features of all the 3D voxels projected into it using mean pooling. This projection allows us to transform an arbitrary number of camera images into a unified BEV frame. Finally, we obtain an image-like BEV feature embedding, which is used to estimate the final representation using a convolutional neural network (CNN) decoder.

This architecture design bridges the image and LiDAR modalities through an intermediate 3D voxelized representation. Hence, we can take LiDAR point clouds as input into the model to directly guide the BEV projection without having to change the overall pipeline. This further enables the distillation of knowledge from the point clouds to images using a teacher-student model.

\subsection{LiDAR-Teacher and Camera-Student} \label{sec:method_teacher}

The co-existence of domain and modality gaps poses additional challenges to the adaptation task. Although the LiDAR sensor in the source domain provides 3D knowledge to the model, it also increases the discrepancy between the two domains, which hurts the model adaptation (as we will see in Sec.~\ref{sec:exp:ablation} and Table~\ref{tab:ablation_modules}). Hence, the unique difficulty of our work lies in exploiting the \textit{LiDAR point clouds} during training to guide the camera model for better 3D estimation.

\noindent \textbf{Depth Supervision by Point Clouds. } The main advantage of point clouds over the image modality is the accurate 3D positional information coming from the depth measurement. Due to the lack of LiDAR during evaluation, we cannot use point clouds as direct input of the model. Hence, one alternative approach to using point clouds is to supervise the depth estimation in the model. As in Eq.~\ref{eq:feature_voxel}, for each pixel, we calculate the features of its corresponding voxels by multiplying the pixel feature with a depth value $\alpha_i$. We use another head to predict a depth distribution $\boldsymbol{\alpha}_\mathrm{px}=\{\alpha_1, \alpha_2, \ldots, \alpha_{N_d}\}$ over $N_d$ depth bins for each pixel $\mathrm{px}$. 

The ground truth depth supervision for this estimation task is generated by LiDAR point clouds as follows: When projected to the image frame, the points corresponding to one pixel can have three conditions. If the pixel has, \textbf{(1)} \textbf{no point inside}: the ground truth depth distribution of it is omitted; \textbf{(2)} \textbf{only one point inside}: the ground truth depth distribution is a one-hot vector, with value one being in the voxel that the point lies in; \textbf{(3)} \textbf{multiple points inside}: the ground truth depth distribution $\alpha_i$ of this point is calculated by counting the number of points in each depth bin, and dividing them by $V_\mathrm{px}$, the total number of points in $\boldsymbol{\alpha}_\mathrm{px}$:
${\alpha}_i = \frac{\text{ Number of points in depth bin } v_i}{\text{ Total number of points in } V_\mathrm{px} }$.

Using a distribution-based depth representation effectively accounts for the ambiguity when objects of different depth occur in one pixel. This happens at the boundary of the objects, and becomes more severe during feature encoding processing when images get down-sampled and each pixel represents larger space. Moreover, a probabilistic depth representation considers uncertainty during depth estimation, and degenerates to pseudo-LiDAR methods~\cite{weng2019monocular} if the one-hot constraint is added.

\vspace{1mm}
\noindent \textbf{Learning from LiDAR-Teacher. } Despite being intuitive and straightforward, direct depth supervision is not optimal for two reasons. First, LiDAR supervision is only on the intermediate feature layer, providing no supervision on the second half of the model. Also, while LiDAR provides accurate depth measurement, ``depth estimation'' is still different from our overall objective on BEV representation. Motivated by this, as shown in Figure~\ref{fig:main}, we propose to use a pretrained LiDAR oracle model to supervise the image model at the final BEV feature embedding, such that the supervision of LiDAR is provided to the whole model and aligns better with the final objective. We call the model using ground truth point cloud information ``LiDAR-Teacher,'' and the model to be supervised ``Camera-Student.'' This boils down to a knowledge distillation problem where the 3D inference knowledge of the LiDAR-teacher is distilled to the camera-only student. Note that the classic problem of ``better teacher, worse student''~\cite{cho2019efficacy,mirzadeh2020improved,zhu2021student} in knowledge distillation due to capacity mismatch does not exist in this model, because the LiDAR-Teacher and Camera-Student models in \ourwork are almost identical.

Overall, this teacher-student mechanism allows the camera model to learn better 3D representation from the point clouds, leading to better LiDAR supervision at different stages, while still keeping the model image-centric for image-only inference.

\begin{table*}[!t]
\vspace{2mm}
\caption{\ourwork leads to significant improvements under \textit{\textbf{day-to-night}} and \textit{\textbf{dry-to-rain}} domain shift settings. Numbers reported in IoU. \textbf{\textit{DA}} and \textbf{\textit{CM}} denote whether a model considers domain adaptation and cross-modality in design, respectively.} \label{tab:d2n-d2r}
\vspace{-1mm}
\begin{minipage}{0.48\linewidth}
\centering
\resizebox{\textwidth}{!}{
\begin{tabu}{@{}l|l|l|l||c|c|c@{}}
\hline
\multicolumn{2}{l|}{\textbf { Day $\rightarrow$ Night }} & {\ {\textit{DA}}\ } & {\ {\textit{CM}}\ } & {\ Vehicle\ } & {\ Road\ } & {\ Lane\ } \\
\hline
\multicolumn{2}{l|}{\text {\ MonoLayout~\cite{mani2020monolayout}}\ \ \ } &  {\ \ \xmark\ }  &  {\ \  \xmark\ }  &  5.9  &  37.7  &  5.9\\
\multicolumn{2}{l|}{\text {\ OFT~\cite{roddick2018orthographic}}\ \ \ }   &  {\ \ \xmark\ }  &  {\ \  \xmark\ }  &  6.6  &  40.5  &  6.0\\ 
\multicolumn{2}{l|}{\text {\ LSS~\cite{philion2020lift}}\ \ \ }           &  {\ \ \xmark\ }  &  {\ \  \xmark\ }  &  6.7  &  41.2  &  7.1\\
\hline
\multicolumn{2}{l|}{\text {\  Wide-range Aug.\ \ }}  &  {\ \ \cmark} &  {\ \  \xmark\ }  &  10.3  &  46.0  &  10.4\\
\multicolumn{2}{l|}{\text {\  Vanilla DA \ \ }}      &  {\ \ \cmark\ }   &  {\ \  \xmark\ }  &  11.2  &  48.8  &  11.1\\
\multicolumn{2}{l|}{\text {\  Depth-Supv DA\ \ }}    &  {\ \ \cmark\ }   &  {\ \  \cmark\ }  &  15.7  &  50.5  &  14.2\\
\multicolumn{2}{l|}{\text {\ Input-fusion Teacher\ \ }}  &  {\ \ \cmark} &  {\ \  \cmark\ }  &  14.9  &  48.8  &  13.1\\
\hline
\multicolumn{2}{l|}{\textbf {\  \ourwork (ours)\ \ }}    &  {\ \  \cmark\ }  &  {\ \  \cmark\ }  &  \textbf{17.0}  &  \textbf{51.8}  &  \textbf{16.9}\\
\hline
\end{tabu}} 
\end{minipage}
\hspace{0.02\linewidth}
\begin{minipage}{0.48\linewidth}
\centering
\resizebox{\textwidth}{!}{
\begin{tabu}{@{}l|l|l|l||c|c|c@{}}
\hline
\multicolumn{2}{l|}{\textbf { Dry $\rightarrow$ Rain }} & {\ {\textit{DA}}\ } & {\ {\textit{CM}}\ } & {\ Vehicle\ } & {\ Road\ } & {\ Lane\ } \\
\hline
\multicolumn{2}{l|}{\text {\ MonoLayout~\cite{mani2020monolayout}}\ \ \ } &  {\ \ \xmark\ }  &  {\ \  \xmark\ }  &  20.6  &  68.7  &  13.1\\
\multicolumn{2}{l|}{\text {\ OFT~\cite{roddick2018orthographic}}\ \ \ }   &  {\ \ \xmark\ }  &  {\ \  \xmark\ }  &  24.1  &  79.8  &  16.2\\ 
\multicolumn{2}{l|}{\text {\ LSS~\cite{philion2020lift}}\ \ \ }           &  {\ \ \xmark\ }  &  {\ \  \xmark\ }  &  27.8  &  71.0  &  16.8\\
\hline
\multicolumn{2}{l|}{\text {\  Wide-range Aug.\ \ }}     &  {\ \ \cmark}  &  {\ \  \xmark\ }  &  28.2  &  71.2  &  17.2\\
\multicolumn{2}{l|}{\text {\  Vanilla DA \ \ }}         &  {\ \ \cmark\ }    &  {\ \  \xmark\ }  &  29.1  &  70.8  &  18.3\\
\multicolumn{2}{l|}{\text {\  Depth-Supv DA\ \ }}       &  {\ \ \cmark\ }    &  {\ \  \cmark\ }  &  \textbf{29.6}  &  71.8  &  19.1\\
\multicolumn{2}{l|}{\text {\ Input-fusion Teacher\ \ }} &  {\ \ \cmark}      &  {\ \  \cmark\ }  &  29.5  &  71.0  &  18.8 \\
\hline
\multicolumn{2}{l|}{\textbf {\  \ourwork (ours)\ \ }}       &  {\ \  \cmark\ }   &  {\ \  \cmark\ }  &  \textbf{29.6}  &  \textbf{71.9}  &  \textbf{19.5}\\
\hline
\end{tabu}} 
\end{minipage}
\vspace{-2mm}
\end{table*}


\begin{table*}[!t]
\begin{minipage}{0.58\linewidth}
\centering
\vspace{2mm}
\caption{\ourwork performs the best under \textit{\textbf{city-to-city}} shift.} \label{tab:b2s}
\vspace{-1mm}
\resizebox{\textwidth}{!}{
\begin{tabu}{@{}l|l|l|l||c|c|c@{}}
\hline
\multicolumn{2}{l|}{\textbf { Boston $\rightarrow$ Singapore }} & {\ {\textit{DA}}\ } & {\ {\textit{CM}}\ } & {\ Vehicle\ } & {\ Road\ } & {\ Lane\ } \\
\hline
\multicolumn{2}{l|}{\text {\ MonoLayout~\cite{mani2020monolayout}}\ \ \ } &  {\ \ \xmark\ }  &  {\ \  \xmark\ }  &  14.2  &  35.9  &  7.5\\
\multicolumn{2}{l|}{\text {\ OFT~\cite{roddick2018orthographic}}\ \ \ }   &  {\ \ \xmark\ }  &  {\ \  \xmark\ }  &  16.8  &  37.9  &  9.6\\ 
\multicolumn{2}{l|}{\text {\ LSS~\cite{philion2020lift}}\ \ \ }           &  {\ \ \xmark\ }  &  {\ \  \xmark\ }  &  17.6  &  38.2  &  10.6 \\
\hline
\multicolumn{2}{l|}{\text {\  Wide-range Aug.\ \ }}     &  {\ \ \cmark}  &  {\ \  \xmark\ }  &  17.9  &  40.5  &  12.4\\
\multicolumn{2}{l|}{\text {\  Vanilla DA \ \ }}         &  {\ \ \cmark\ }    &  {\ \  \xmark\ }  &  13.0  &  31.4  &  9.1 \\
\multicolumn{2}{l|}{\text {\  Depth-Supv DA\ \ }}       &  {\ \ \cmark\ }    &  {\ \  \cmark\ }  &  19.0  &  42.8  &  14.9\\
\multicolumn{2}{l|}{\text {\ Input-fusion Teacher \ \ }}  &  {\ \ \cmark\ }  &  {\ \  \cmark\ }  &  18.6  &  42.7  &  14.1\\
\hline
\multicolumn{2}{l|}{\textbf {\  \ourwork (ours)\ \ }}       &  {\ \  \cmark\ }   &  {\ \  \cmark\ }  &  \textbf{20.5}  &  \textbf{43.1}  &  \textbf{15.6} \\
\hline
\end{tabu}} 
\end{minipage}
\hspace{0.02\linewidth}
\begin{minipage}{0.38\linewidth}
\centering
\vspace{2mm}
\caption{\ourwork performs the best under \textit{\textbf{dataset-to-dataset}} domain gaps in IoU.} \label{tab:n2l}
\vspace{-1mm}
\resizebox{\textwidth}{!}{
\begin{tabu}{@{}l|l|l|l||c@{}}
\hline
\multicolumn{2}{l|}{\textbf { nuScenes $\rightarrow$ Lyft }} & {\ {\textit{DA}}\ } & {\ {\textit{CM}}\ } & {\ Vehicle\ } \\
\hline
\multicolumn{2}{l|}{\text {\ MonoLayout~\cite{mani2020monolayout}}\ \ \ } &  {\ \ \xmark\ }  &  {\ \  \xmark\ }  &  11.8    \\
\multicolumn{2}{l|}{\text {\ OFT~\cite{roddick2018orthographic}}\ \ \ }   &  {\ \ \xmark\ }  &  {\ \  \xmark\ }  &  16.5\\ 
\multicolumn{2}{l|}{\text {\ LSS~\cite{philion2020lift}}\ \ \ }           &  {\ \ \xmark\ }  &  {\ \  \xmark\ }  &  19.9  \\
\hline
\multicolumn{2}{l|}{\text {\  Wide-range Aug.\ \ }}                      &  {\ \ \cmark$^*$}  &  {\ \  \xmark\ }  &  21.9    \\
\multicolumn{2}{l|}{\text {\  Vanilla DA \ \ }}                          &  {\ \ \cmark\ }  &  {\ \  \xmark\ }  &  22.5     \\
\multicolumn{2}{l|}{\text {\  Depth-Supv DA\ \ }}                  &  {\ \ \cmark\ }  &  {\ \  \cmark\ }  &  23.4  \\
\multicolumn{2}{l|}{\text {\ Input-fusion Teacher \ \ }}  &  {\ \ \cmark}  &   {\ \  \cmark\ }  &  22.8   \\
\hline
\multicolumn{2}{l|}{\textbf {\  \ourwork (ours)\ \ }}                        &  {\ \  \cmark\ }  &  {\ \  \cmark\ }  &  \textbf{24.4} \\
\hline
\end{tabu}} 
\end{minipage}
\vspace{-4mm}
\end{table*}


\subsection{Cross-Domain Adaptation}\label{sec:method_adv}

Since the BEV annotations and the LiDAR ground truth are only available in the source data, the model will be heavily biased to the source distribution during teacher-student supervision. Hence, we bridge the target and source domains using adversarial training. Specifically, we place one discriminator $D_1$ at the BEV decoder CNN blocks, and another $D_2$ at the image encoder CNN blocks, to align the features of two domains by optimizing over discriminator losses. While the final-layer discriminator $D_1$ is constantly useful to align features learned from the LiDAR-Teacher and final ground truth, we find that the middle-layer discriminator $D_2$ is very effective under certain domain gaps where images have great changes but LiDAR remains robust. 

To achieve adversarial learning, given a feature encoder $E$ and input sample $X$, a domain discriminator $D$ is used to discriminate whether the feature $E(X)$ comes from the source domain or the target domain. The target and source domain samples are given the label $d=1$ and $d=0$, respectively. And $D(E(X))$ outputs the probability of the sample $X$ belonging to the target domain. Hence, the discriminator loss is formulated by a cross-entropy loss:

{\small
    \vspace{-3mm}
    \setlength{\abovedisplayskip}{5.5pt}
    \setlength{\belowdisplayskip}{5.5pt}
    \setlength{\abovedisplayshortskip}{0pt}
    \setlength{\belowdisplayshortskip}{3pt}
\begin{align}\label{eq:loss_dis}
    \mathcal{L}_{\mathrm{dis}}=d \log D(E(X))+(1-d) \log (1-D(E(X))).
\end{align}}%
Moreover, in order to learn domain-invariant features, our feature encoder $E$ should try to extract features that fool the discriminator $D$, while the discriminator $D$ tries to distinguish the right domain label of the samples. This adversarial strategy can be formulated as a ``min-max'' optimization problem: $\mathcal{L}_{\mathrm{D}}=\min _{E} \max _{D} \mathcal{L}_{\mathrm{dis}}$. The ``min-max'' problem is achieved by a Gradient Reverse Layer (GRL)~\cite{ganin2015grl}, which produces reverse gradient from the discriminator $D$ to learn the domain-invariant encoder $E$. The loss form is the same for both $D_1$ and $D_2$.

\subsection{Full Objective and Inference}\label{sec:method_loss}

The overall objective of our model is composed of the supervision from the BEV ground truth, the LiDAR-Teacher, and the domain alignment discriminators. Given the output rasterized BEV representation map $\boldsymbol{y}\in \mathbb{R}^{X\times Y \times C}$, the ground truth (GT) loss term $\mathcal{L}_{\mathrm{GT}}$ can be formulated as a cross-entropy loss between the estimated source domain BEV map $\boldsymbol{\tilde y}^s$ and the GT label $\boldsymbol{y}^s$:
{\small
    \vspace{-3mm}
    \setlength{\abovedisplayskip}{4pt}
    \setlength{\belowdisplayskip}{5pt}
    \setlength{\abovedisplayshortskip}{0pt}
    \setlength{\belowdisplayshortskip}{3pt}
\begin{align}\label{eq:loss_final_obj}
    \mathcal{L}_{\mathrm{GT}}(\boldsymbol{\tilde y}^s,\boldsymbol{y}^s) = -\sum_{i=1}^X\sum_{j=i}^Y\sum_{k=1}^C y_{(i,j,k)}^s \log \tilde y_{(i,j,k)}^s.
\end{align}}%
The supervision from the LiDAR-Teacher is composed of a direct depth estimation loss $\mathcal{L}_{\mathrm{dp}}$ and a teacher feature supervision $\mathcal{L}_{\mathrm{T}}$. As described in Sec.~\ref{sec:method_bev}, given the 3D depth volume $\boldsymbol{\alpha}\in \mathbb{R}^{H\times W \times N_d}$, the direct depth supervision term $\mathcal{L}_{\mathrm{dp}}$ is formulated as a cross-entropy loss between the estimated 3D depth distribution volume $\boldsymbol{\tilde \alpha}^s$ in the source domain, and the GT depth volume $\boldsymbol{\alpha}^s$ calculated from LiDAR point clouds as described in Sec.~\ref{sec:method_teacher}:

{\small
    \vspace{-3mm}    
    \setlength{\abovedisplayskip}{3pt}
    \setlength{\belowdisplayskip}{4pt}
    \setlength{\abovedisplayshortskip}{0pt}
    \setlength{\belowdisplayshortskip}{3pt}
\begin{align}\label{eq:loss_depth}
    \mathcal{L}_{\mathrm{dp}}(\boldsymbol{\tilde \alpha}^s,\boldsymbol{\alpha}^s) = -\sum_{i=1}^H\sum_{j=i}^W\sum_{k=1}^{N_d} \alpha_{(i,j,k)}^s \log \tilde \alpha_{(i,j,k)}^s.
\end{align}}%
And for the LiDAR-Teacher feature supervision: $\mathcal{L}_{\mathrm{T}}(\boldsymbol{F}^{\mathrm{te}},\boldsymbol{F}^{\mathrm{st}}) = \mathcal L_2 (\boldsymbol{F}^{\mathrm{te}},\boldsymbol{F}^{\mathrm{st}})$ is an $\mathcal{L}_{\mathrm{2}}$ loss, where $\boldsymbol{F}^{\mathrm{te}}$ and $\boldsymbol{F}^{\mathrm{st}}$ are the feature maps of teacher and student models, respectively. Finally, the domain adaptation loss contains $\mathcal{L}_{\mathrm{D_1}}$ and $\mathcal{L}_{\mathrm{D_2}}$ with the form described in Eq.~\ref{eq:loss_dis}.

\vspace{2mm}
\noindent \textbf{The final objective } is formulated as a multi-task optimization problem:
{
\small
\begin{align}\label{eq:loss_overall}
    \mathcal{L}_{\mathrm{\ourwork}} = \mathcal{L}_{\mathrm{F}} + \mathcal{\lambda}_{\mathrm{T}}\mathcal{L}_{\mathrm{T}} + \mathcal{\lambda}_{\mathrm{dp}}\mathcal{L}_{\mathrm{dp}} + \mathcal{\lambda}_{\mathrm{D_1}}\mathcal{L}_{\mathrm{D_1}} + \mathcal{\lambda}_{\mathrm{D_2}}\mathcal{L}_{\mathrm{D_2}},
\end{align}}%
where $\mathcal{\lambda}_{\mathrm{T}}, \mathcal{\lambda}_{\mathrm{dp}}, \mathcal{\lambda}_{\mathrm{D_1}}, \text{and } \mathcal{\lambda}_{\mathrm{D_2}}$ are weights for the corresponding loss terms. The \ourwork model is trained end-to-end using the loss term in Eq.~\ref{eq:loss_overall}. During inference, target samples are fed into the Camera-Student model to output the final BEV representation. More training details are provided in Sec.~\ref{sec:exp}.

\section{Experiments} \label{sec:exp}

\noindent\textbf{Datasets and Domain Settings.} We evaluate DualCross with four unique domain shift settings constructed from two large-scale datasets, nuScenes~\cite{Caesar2019nuScenes} and Lyft~\cite{lyftdataset}, following existing LiDAR-based domain adaptation work, including SRDAN~\cite{zhang2021srdan}, ST3D~\cite{yang2021st3d}, UDA3D~\cite{luo2021unsupervised}, and xMUDA~\cite{jaritz2020xmuda}. Specifically, for the \textit{day-to-night}, \textit{city-to-city}, and \textit{dry-to-rain} settings, we use the sentence in the nuScenes dataset and filter the keywords to split the dataset into corresponding subsets to create the intra-class adaptation scenarios. For the \textit{dataset-to-dataset} setting, we use the official split of the nuScenes dataset, and the split provided in ST3D~\cite{yang2021st3d} for the Lyft dataset. All adaptation settings follow the assumption that the source has access to cameras and LiDAR sensors, while the target only has cameras. We use all six cameras provided by the nuScenes dataset. We also analyze surprising observations on cross-modality performance in the ablation study.

\vspace{2mm}\noindent\textbf{Implementation Details.} Following \cite{philion2020lift}, we use EfficientNet~\cite{tan2019efficientnet} pretrained on ImageNet as our image encoder backbone. Two heads are applied to estimate pixel features and pixel-wise depth distribution from the $8\times$ down-sampled feature map. The 3D feature maps are projected to the BEV frame using mean pooling. For the BEV decoder we use ResNet-18 as the backbone, and upsample the features learned from the first three meta-layers of ResNet to the final BEV output. The $D_1$ and $D_2$ domain discriminators are applied to the output feature layers of EfficientNet and ResNet backbones, respectively. We use a light-weight discriminator architecture, which is composed of a global averaging pooling layer, followed by two fully-connected layers, and outputs the domain label. For input, we resize and crop input images to size $128 \times 352$. For output, we consider a $100 \text { meters} \times 100 \text { meters}$ range centered at the ego-vehicle, with the grid size set to be $0.5 \text { meters} \times 0.5 \text { meters}$. The depth bin is set to be $1.0 \text { meter}$ between $4.0 \text { meters}$ and $45.0 \text{ meters}$ range. The whole model is trained end-to-end, with $\lambda_{\mathrm{T}}=1.0, \lambda_\mathrm{dp}=0.05, \lambda_\mathrm{D_1}=0.1, \lambda_\mathrm{D_2}=0.01$. We train DualCross using the Adam~\cite{kb14adam} optimizer with learning rate $0.001$ and weight decay 1$e$-7 for $50$K steps for the teacher model, and $200$K for the student model. We use horizontal flipping, random cropping, rotation, and color jittering augmentation during training. The whole model is implemented using the PyTorch framework~\cite{paszke2019pytorch}.

\begin{figure*}[!t]
    \centering
    \vspace{2mm}
    \includegraphics[trim=49 34 49 58,clip, width=0.98\linewidth]{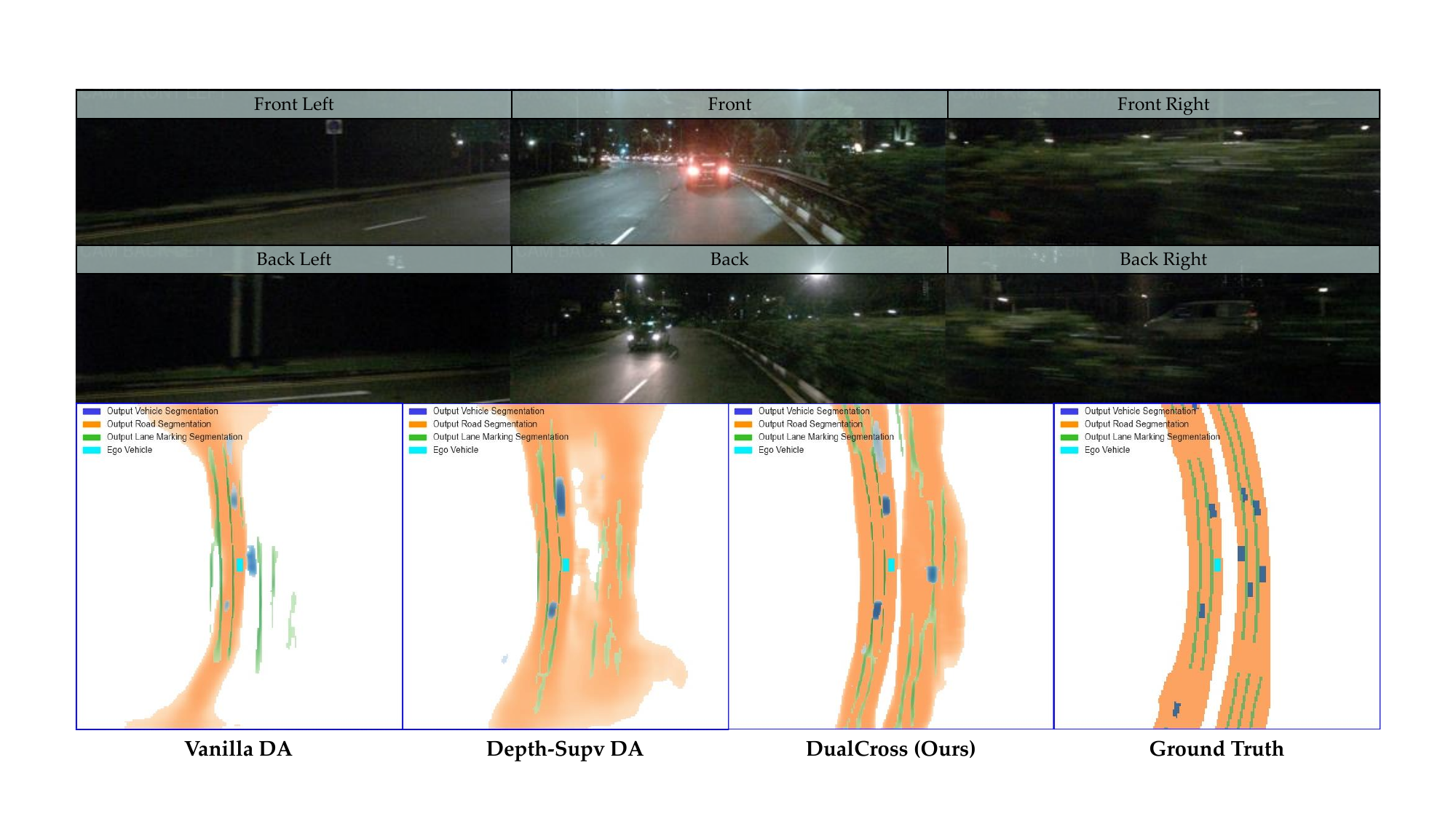}
    \vspace{-2mm}
    \caption{\textbf{Qualitative Results in Day $\mathbf{\rightarrow}$ Night setting} (model is trained with daytime data, and validated with night data). We notice that DualCross performs significantly better than other baselines for \textcolor{myblue}{vehicles}, \textcolor{myorange}{drivable roads}, and \textcolor{mygreen}{lane marking} classes. From \textbf{left} to \textbf{right}: (1) Vanilla adversarial learning; (2) LiDAR as depth supervision with adversarial learning; (3) our \ourwork model; (4) Ground Truth. Best viewed in color.}
    \vspace{-5mm}
    \label{fig:main_results}
\end{figure*}

\subsection{BEV Segmentation Results and Comparisons}

\noindent \textbf{Baselines. } We compare our method with state-of-the-art BEV 3D layout perception work MonoLayout~\cite{mani2020monolayout}, OFT~\cite{roddick2018orthographic}, LSS~\cite{philion2020lift}, as well as other baseline methods in domain adaptation and cross-modality learning. \textit{Wide-range Aug.} means using a wide range of random scaling augmentation which potentially includes the target domain scale. For \textit{Vanilla DA}, we adapt camera-only DA-Faster~\cite{chen2018dafaster} to our BEV perception setting. \textit{Depth-Supv DA} stands for depth-supervised domain adaptation. We use source domain LiDAR as ground truth to supervise the depth estimation during training, without LiDAR-Teacher supervision (only $\mathcal{L}_\mathrm{dp}$ without $\mathcal{L}_\mathrm{T}$). \textit{Input-fusion Teacher} is an alternative way of designing the LiDAR-Teacher, where we directly fuse point $(x,y,z)$ coordinates into their corresponding image pixels as additional channels in the teacher model, similar to Pointpainting~\cite{vlh20pointpainting}. We use \textbf{\textit{DA}} and \textbf{\textit{CM}} to denote whether a model considers domain adaptation and cross-modality in design, respectively. Results are reported on vehicle, drivable roads, and lane marking classes using intersection-over-union (IoU).

\vspace{2mm}
\noindent \textbf{Day-to-Night Adaptation. } As shown in Table~\ref{tab:d2n-d2r} on the left, we observe that our \ourwork model achieves the best performance on all classes. We notice that the improvement under the Day $\rightarrow$ Night setting is exceptionally high. This is because the initial domain gap between day and night scenarios is very large in the camera modality space. Moreover, the LiDAR sensor is robust under illumination changes, due to its active imaging mechanism as opposed to camera's passive one. Thus, incorporating LiDAR point cloud information helps the model to learn a more robust, illumination-invariant representation from the image inputs.


\vspace{2mm}
\noindent \textbf{Dry-to-Rain Adaptation. } As shown in Table~\ref{tab:d2n-d2r} on the right, under this setting we also observe that our \ourwork model achieves the best performance on all classes. We notice that the improvement under the Dry $\rightarrow$ Rain setting is not as big as the previous setting. This is because the domain gap between dry and rain scenarios is not big in the image modality. Hence, baseline methods OFT and LSS are already able to obtain decent results even without domain adaptation. Furthermore, rainy weather is known to cause great domain shift in the LiDAR modality~\cite{xu2021spg}. As a result, the knowledge learned from source-exclusive LiDAR suffers from an unknown domain shift which hinders its usefulness. This can potentially cancel out the benefit of 3D information learned from point clouds and explains for the smaller improvement.

\vspace{2mm}
\noindent \textbf{Dataset-to-Dataset Adaptation. } As shown in Table~\ref{tab:n2l}, we also observe that our \ourwork model achieves the best performance in the nuScenes $\rightarrow$ Lyft setting. Following \cite{philion2020lift}, because Lyft does not provide road segment and lane marking information in the HD map, we report results on the vehicle class. Compared with baselines with and without domain adaptation or cross-modality learning, our \ourwork demonstrates superior performance in leveraging and adapting LiDAR information.

\vspace{2mm}
\noindent \textbf{City-to-City Adaptation. } As shown in Table~\ref{tab:b2s}, we observe that our \ourwork model achieves the best performance on all classes for two inter-city transfer settings. Without domain adaptation, baseline approaches MonoLayout, OFT, and LSS all suffer from performance degradation. Direct depth supervision and alternative input-fusion teacher models do not bring as much improvement as \ourwork. The results clearly demonstrate the effectiveness of our method by distilling and aligning the LiDAR information for cross-modality 3D BEV perception.

\vspace{2mm}
\noindent \textbf{Qualitative Results. } As shown in Figure~\ref{fig:main_results}, under the Day $\mathbf{\rightarrow}$ Night domain shift setting, our model achieves significantly better monocular 3D perception than other baselines. We observe that \ourwork provides more clearly defined road boundaries and lane markings. The depth and size of the vehicles and the road on the right side are also predicted more accurately. \ourwork only misses some vehicles that are hardly visible in camera due to occlusion and distance. Overall, the qualitative results validate the effectiveness of \ourwork in closing the gap between data domains and leveraging point clouds information for better 3D inference. 

\begin{table}[!t]
\caption{Our proposed components all contribute to the final performance. We report results on the vehicle class under the \textbf{\textit{day-to-night}} domain gap in IoU. \textit{WA, AD, LS, LT} stand for Wide Augmentation, Adversarial Discriminators, LiDAR Supervision, and LiDAR-Teacher, respectively.} \label{tab:ablation_modules}
\vspace{-1mm}
\centering
\resizebox{0.42\textwidth}{!}{
\begin{tabu}{@{}c|c|c|c|c||c|c@{}}
\hline
 {\ \ Baseline} & WA & AD & LS & LT & {Results} & \textit{diff}\\
\hline
\cmark  &    &    &    &      &      6.7 & $0$ \\\hline
\cmark  &    &    &  \cmark   &    & 6.4 & \textcolor{red}{$-0.3$} \\\hline
\cmark  &  \cmark  &    &     &    &  10.3 & \textcolor{mygreen}{$+3.6$} \\\hline
\cmark  &  \cmark  &  \cmark  &    &    &  11.2  & \textcolor{mygreen}{$+4.5$}\\\hline
\cmark  &  \cmark  &  \cmark  &  \cmark  &    &  15.7 & \textcolor{mygreen}{$+9.0$}\\\hline
\cmark  &  \cmark  &  \cmark  &  \cmark  &  \cmark  & \textbf{17.0} & \textcolor{mygreen}{$+10.3$} \\
\hline
\end{tabu}} 
\vspace{-2mm}
\end{table}


\subsection{Analysis and Ablation Study}\label{sec:exp:ablation}

\noindent \textbf{Direct LiDAR Supervision Leads to Worse Performance.} It is naturally believed that introducing multiple sensors in the perception model is bound to increase the model performance. Surprisingly, experiments shown in Table~\ref{tab:ablation_modules} negate this naive intuition. When we introduce the LiDAR sensor in the source domain as depth supervision, the result decreases by 0.3. As we described in Sec.~\ref{sec:method_teacher}, the domain distribution divergence increases after introducing the sensor-modality shift. As a result, we propose multiple components in \ourwork to account for the visual and sensor domain shifts. Experiments show that while the wide augmentation strategy and adversarial discriminator both achieve better results than the baseline $(11.2 $ vs. $6.7$ in IoU$)$, our LiDAR-Teacher further boosts the result to $17.0$ by leveraging effective LiDAR knowledge distillation.

\begin{figure}[!t]
\begin{minipage}{0.95\linewidth}
\vspace{4pt}
\includegraphics[trim=210 128 210 128,clip, width=\linewidth]{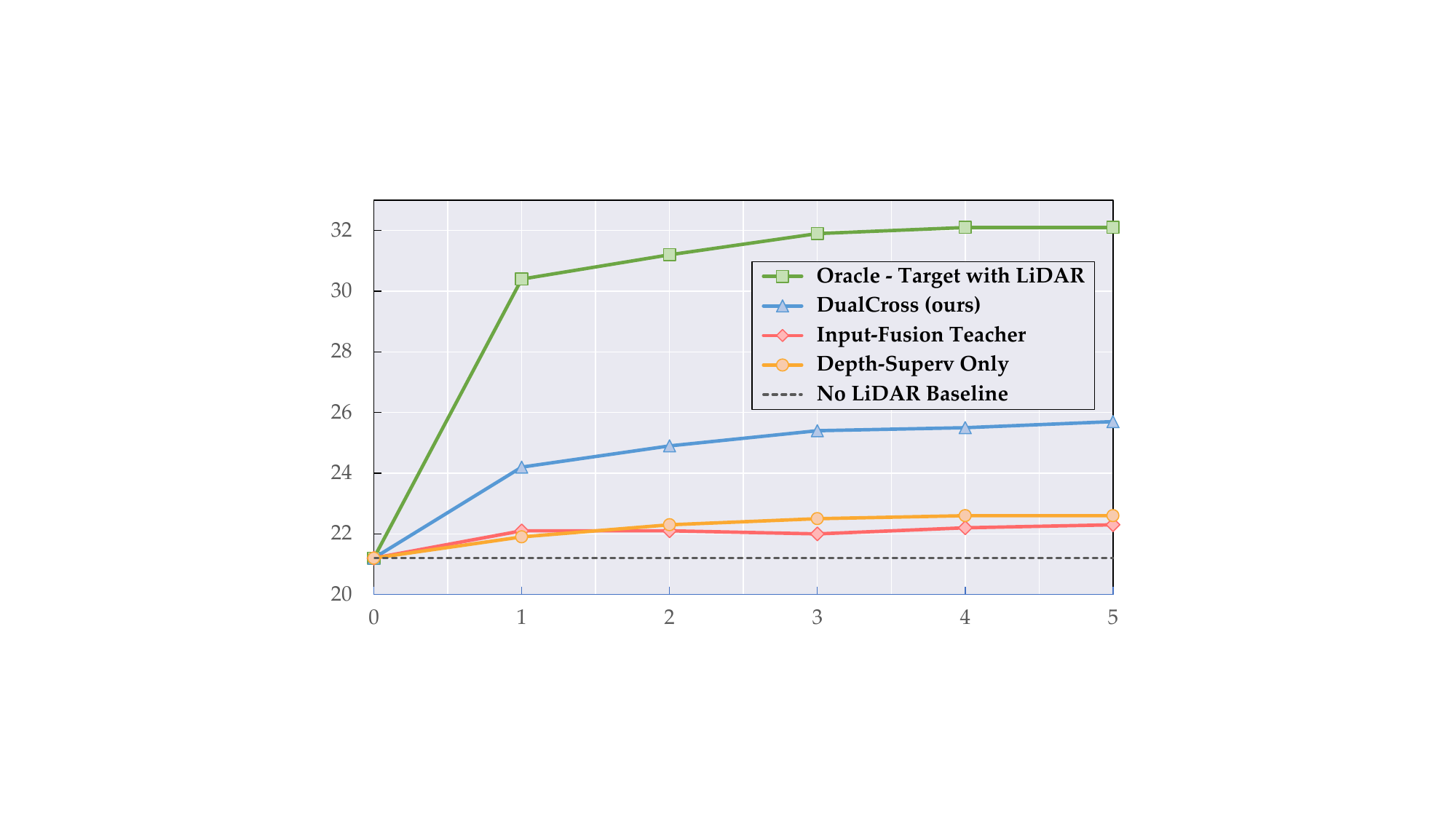}
\vspace{-6mm}
\caption{Results of \ourwork improve as the number of LiDAR points increases.}
\label{fig:ablation-lidar-sweep}
\end{minipage}
\vspace{-5mm}
\end{figure}

\vspace{1mm}
\noindent \textbf{LiDAR Density \& Comparison with Oracle Model. } As shown in Figure~\ref{fig:ablation-lidar-sweep}, we validate that our model achieves higher performance when denser LiDAR is available. This can be accomplished by grouping continuous scans of LiDAR point clouds (from 1 to 5) into a single unit, to have a denser 3D representation of the scene. We observe that other cross-modality baselines including \textit{Input-Fusion Teacher} and \textit{Depth-Supv} models cannot effectively leverage the LiDAR knowledge, even with dense point clouds available. We also compare our model with the LiDAR oracle model (target domain also has the LiDAR modality) and find that the gap between the upper bound result and the No-LiDAR baseline is significantly reduced. The remaining performance gap is caused by the unknown LiDAR domain gap which we hope to further reduce in future work.

\vspace{1mm}
\noindent \textbf{Dealing with Mixed Domain Shift. } Another common but under-explored question we observe in the 3D domain adaption setting is the mixed domain shift problem, where multiple types of gaps between source and target domains occur concurrently. For example, in the nuScenes dataset, the Boston data are collected exclusively during daytime, whereas the Singapore data encompass both day and night captures. This leads to a mixture of city-wise and lighting-wise domain shifts. As shown in Table~\ref{tab:ablation_domain_mixed}, we find that directly leveraging domain adaptation in this scenario leads to worse performance than direct inference, because mixed domains in the target confuse the discriminator. Hence, we propose a progressive learning mechanism, where we first perform adaptation with city-wise data for $100K$ steps, and then train the model on the full target domain dataset for another $150K$ steps. This effectively alleviates the mixed domain shift problem, and helps \ourwork achieve leading results than other baselines.

\vspace{1mm}
\noindent \textbf{Computational Complexity. } Table~\ref{tab:complexity} summarizes the number of parameters and inference speed for prior baselines and our model. Our Lidar-Teacher distillation and multi-level adversarial learning modules do not affect the inference efficiency of DualCross compared with the baseline. Our total number of parameters is 15M, and our inference time is 33 Frame-per-Second (FPS) on a V100 GPU, which is on par with the baseline LSS~\cite{philion2020lift}. The training time for our model is around 20 hours on 4$\times$V100 GPUs.

\begin{table}[!t]
\begin{minipage}{0.98\linewidth}
\vspace{2mm}
\caption{The proposed progressive learning strategy effectively addresses the challenge caused by the mixed domain gap scenario (\textbf{\textit{Boston-to-Singapore}} mixed with \textbf{\textit{day-to-night}}) on nuScenes.} \label{tab:ablation_domain_mixed}
\vspace{-1mm}
\centering
\resizebox{0.95\textwidth}{!}{
\begin{tabu}{@{}l|l|c|c|c@{}}
\hline
\multicolumn{2}{l|}{\textbf { Mixed Domain Gap }} & {\ Vehicle\ } & {\ Road\ } & {\ Lane\ } \\
\hline
\multicolumn{2}{l|}{\text { Direct Inference }}  &   17.6  &  38.2  &  10.6   \\
\multicolumn{2}{l|}{\text { Vanilla DA }}        &   13.0  &  31.4  &  9.1   \\
\multicolumn{2}{l|}{\text { Progressive DA }}     &   18.8  &  41.5  & 13.2    \\
\multicolumn{2}{l|}{\textbf { \ourwork (ours)}}      &   \textbf{20.5}  &  \textbf{43.1}  &  \textbf{15.6}   \\
\hline
\end{tabu}} 
\end{minipage}
\vspace{-2mm}
\end{table}


\begin{table}[!t]
\caption{\ourwork achieves great perception results with efficient inference time compared with the baselines.} \label{tab:complexity}
\vspace{-1mm}
\centering
\resizebox{0.48\textwidth}{!}{
\begin{tabu}{@{\hspace{2mm}}lcc@{\hspace{2mm}}}
\toprule
 & \#Params (M) & Frame-per-Second (FPS)\\
\midrule
OFT~\cite{roddick2018orthographic} & 22   &  25\\ 
LSS~\cite{philion2020lift} & 14   &  35   \\ 
\textbf{DualCross (Ours)} & 15   &  33\\ 
\bottomrule
\end{tabu}} 
\vspace{-4mm}
\end{table}

\section{Conclusion}

In this paper, we proposed \ourwork to estimate 3D scene representation in BEV under domain shift and modality change. To achieve this, we construct a LiDAR-Teacher and distill knowledge from it into a Camera-Student by feature supervision. And we further propose to align feature space between the domains using multi-stage adversarial learning. Results on large-scale datasets with various challenging domain gaps demonstrated the effectiveness of our approach, which marks a significant step towards robust 3D scene perception in the wild.


{\small
\bibliographystyle{ieee_fullname}
\bibliography{egbib}
}

\clearpage

\appendix

\section{Dataset Details} \label{sec:supp-dataset}


In this section, we explain our dataset split in more details. For our experiments, we always split the target domain into two subsets for a fair comparison. One of them can be accessed during training for adversarial learning and domain alignment, and the other is held out exclusively for validation.

\vspace{2mm}
\noindent \textbf{City-to-City Adaptation. } For the \textit{city-to-city} adaptation scenario, we sub-sample the trainval split of the large-scale dataset nuScenes~\cite{Caesar2019nuScenes} captured in Boston and Singapore cities. We treat one city as the source domain and the other as the target domain. The Boston part of data has 467 scenes in total, which is separated into 350 scenes for training and 117 scenes for validation. And the Singapore part of data has 383 scenes in total, which is separated into 287 scenes for training and 96 scenes for validation.

\vspace{2mm}
\noindent \textbf{Day-to-Night Adaptation. } For the \textit{day-to-night} adaptation scenario, we also sub-sample the trainval split of the large-scale dataset nuScenes~\cite{Caesar2019nuScenes}. Every scene in the nuScenes dataset has a sentence of description, which can be parsed and used to categorize it into certain class. In this way, we create a day scene subset and a night scene subset out of the whole dataset. We treat day as the source domain and night as the target domain, because the night subset has significantly fewer samples than the day subset. The day split has 751 scenes which are all used for training. And the night part has 99 scenes in total, which is separated into 74 scenes for training and 25 scenes for validation.

\vspace{2mm}
\noindent \textbf{Dry-to-Rain Adaptation. } For the \textit{dry-to-rain} adaptation scenario, we also sub-sample the trainval split of the large-scale dataset nuScenes~\cite{Caesar2019nuScenes}. Similarly, using the scene description sentence, we create a dry (non-rainy) scene subset and a rainy scene subset out of the whole dataset. We treat dry as the source domain and rain as the target domain, because the rain subset has significantly fewer samples than the dry subset. The dry split has 685 scenes which are all used for training. And the rain part of data has 165 scenes in total, which is separated into 124 scenes for training and 41 scenes for validation.

\vspace{2mm}
\noindent \textbf{Dataset-to-Dataset Adaptation. } For the \textit{dataset-to-dataset} adaptation scenario, we use the two large-scale autonomous driving datasets nuScenes~\cite{Caesar2019nuScenes} and Lyft~\cite{lyftdataset}. We treat one dataset as the source domain and the other as the target domain. For the nuScenes dataset, we use the original train and validation split, which has 700 scenes and 150 scenes, respectively. The Lyft dataset does not have an original split, so we sub-sample 132 scenes for training and 48 scenes for validation.

\section{Additional Results and Analysis}
\label{sec:supp-exp}

\subsection{Unique Challenge Due to Problem Formulation } 

We provide another perspective to understand the challenges brought by the co-existence of cross-modality and cross-domain gaps, thus further motivating the design of our architecture. It is proved that in domain adaptation, the target domain error $\epsilon^{t}(h)$ for input $h$ can be upper-bounded by the inequality~\cite{ben2010theory}:

{
    \setlength{\abovedisplayskip}{-3.0pt}
    \setlength{\belowdisplayskip}{6.0pt}
    \setlength{\abovedisplayshortskip}{0pt}
    \setlength{\belowdisplayshortskip}{0pt}
\begin{align}
    \epsilon^{t}(h) \leq \epsilon^{s}(h)+d_{\mathcal{H} \Delta \mathcal{H}}\left(P_{X}^{s}, P_{X}^{t}\right)+C,
\end{align}}%
where the bound is composed of the source-domain error $\epsilon^{s}(h)$, source-target distribution divergence $d_{\mathcal{H} \Delta \mathcal{H}}\left(P_{X}^{s}, P_{X}^{t}\right)$, and another term which is considered constant in our case. Existing domain adaptation work mostly focuses on reducing the source-domain error. However, while the introduction of the LiDAR sensor \textbf{reduces the first term}, it \textbf{increases the second term}, because the source and target domains have an additional sensor-type gap. It is also demonstrated in Sec.~\ref{sec:exp} that naively using source domain LiDAR can even hinder the target performance rather than improve it. Hence, the unique difficulty of our work lies in \textit{leveraging LiDAR to reduce the source-domain error $\epsilon^{s}(h)$ and, in the meanwhile, preventing distribution divergence $d_{\mathcal{H} \Delta \mathcal{H}}\left(P_{X}^{s}, P_{X}^{t}\right)$ from increasing too much}. 

\begin{table*}[!t]
\caption{We validate the design choices of \ourwork by comparing with various depth GT generation methods and different choices of teacher supervision signals. Results show that depth supervision generated by Point Number Distribution (Eq.~\ref{eq:feature_voxel} in the main paper), and feature-level supervision from LiDAR-Teacher helps the model achieve the best performance.} \label{tab:depth-supervision}
\vspace{0.05cm}
\centering
\resizebox{0.8\textwidth}{!}{
\begin{tabu}{@{}c|l|c|c|c@{}}
\hline
\multicolumn{2}{l|}{\textbf { Design choices in Day $\rightarrow$ Night Setting }} & {\ Vehicle\ } & {\ Road\ } & {\ Lane\ } \\
\hline\hline
\multirow{5}{*}{ \shortstack[c]{ Depth Ground Truth Signal \\ (From Point Clouds) }} & {\text { No LiDAR }} &  11.2  &  48.8  &  11.1\\
& {\text { Majority Voting }}   &  14.1  &  50.0  &  13.7\\ 
& {\text { Random Selection }}   &  13.8  &  49.5  &  12.6\\ 
& {\text { Softmax Point Number Distribution}\ \ \ }          &  13.1  &  49.9  &  11.9\\
& {\textbf { Point Number Distribution (Ours)}\ \ \ }           &  \textbf{17.0}  &  \textbf{51.8}  &  \textbf{16.9}\\
\hline\hline
\multirow{3}{*}{ \text { LiDAR-Teacher Supervision } } & {\text { No LiDAR-Teacher }} &  15.7  &  50.5  &  14.2\\
& {\text { Soft-label Supervision }}   &  15.2  &  50.7  &  13.9\\ 
& {\textbf { Feature Supervision (Ours) }}   &  \textbf{17.0}  &  \textbf{51.8}  &  \textbf{16.9}\\ 
\hline
\end{tabu}} 
\vspace{-2mm}
\end{table*}


\subsection{Different Depth GT Signal Generation Methods } As shown in Table~\ref{tab:depth-supervision}, we validate \ourwork's design choice in depth ground truth signal generation (Eq.~\ref{eq:feature_voxel} in the main paper) by comparing with different alternative methods. Note that all the methods have the same output when there is no point or only one point projected to a pixel. The difference only comes from the behaviour when multiple points are projected to one pixel. 
\begin{itemize}
    \item \textit{Majority Voting} means to generate a one-hot GT vector by assigning ``1'' to the depth bin with the most number of points inside, and randomly select one bin when more than one bin with the most number of points.
    \item \textit{Random Selection} means to generate a one-hot GT vector by randomly selecting one of all the projected points, and assigning ``1'' to the depth bin in which the selected point lies.
    \item \textit{Softmax Point Number Distribution} means to count the number of points inside every depth bin, and then use softmax to turn this vector into a distribution.
\end{itemize}
Finally, in \ourwork we use direct \textit{Point Number Distribution}, where we count the number of points inside every depth bin, and divide the vector by the total number of points to get a depth distribution. Results show that our direct point number distribution outperforms other counterparts in ground truth generation. One possible reason is because \ourwork down-samples the feature map $8\times$ when estimating the 3D depth volume. This makes it much more common for multiple points with different depth values to fall in the same pixel. By contrast, the other two methods, \textit{Majority Voting} and \textit{Random Selection}, will lose valuable information in these cases. On the other hand, softmax activation over-smooths the distribution, and also assigns small values to zero bins (depth bins with no point inside). Hence, we find that preserving the native depth distribution of the points is better to supervise \ourwork in 3D evaluation.

\subsection{Different LiDAR-Teacher Supervision Design} 

As shown in Table~\ref{tab:depth-supervision}, we also validate \ourwork's design choice of LiDAR-Teacher Supervision by comparing with different alternative methods. In addition to the feature-level supervision used in \ourwork, another commonly used teacher supervision is the soft label output. Specifically, \textit{Soft-label Supervision} means to use the class distribution output of the teacher model as the supervision of the student model, as opposed to the one-hot vector from the ground truth annotation. We find that the feature-level supervision performs better than the soft-label supervision. One reason is because we use a small number of classes, which makes the supervision from the soft labels less informative. Moreover, because the teacher and student models in \ourwork have almost identical architecture and capacity, enforcing the corresponding feature level similarity between the teacher and student models provides a stronger supervision than the soft-label output, without harming the model learning. As future work, we will explore whether soft labels may be more useful when the number of classes in the model is larger.

\begin{table}[!t]
\caption{\ourwork achieves the best performance on all classes under \textit{\textbf{Lyft-to-nuScenes}} domain gaps in IoU. {\textit{DA}} and {\textit{CM}} mean domain adaptation and cross modality, respectively.} \label{tab:supp-l2n}
\vspace{0.05cm}
\centering
\resizebox{0.43\textwidth}{!}{
\begin{tabu}{@{}l|l|l|l||c@{}}
\hline
\multicolumn{2}{l|}{\textbf { Lyft $\rightarrow$ nuScenes }} & {\ {\textit{DA}}\ } & {\ {\textit{CM}}\ } & {\ Vehicle\ } \\
\hline
\multicolumn{2}{l|}{\text {\ MonoLayout~\cite{mani2020monolayout}}\ \ \ } &  {\ \ \xmark\ }  &  {\ \  \xmark\ }  &  7.1    \\
\multicolumn{2}{l|}{\text {\ OFT~\cite{roddick2018orthographic}}\ \ \ }   &  {\ \ \xmark\ }  &  {\ \  \xmark\ }  &  11.9    \\ 
\multicolumn{2}{l|}{\text {\ LSS~\cite{philion2020lift}}\ \ \ }           &  {\ \ \xmark\ }  &  {\ \  \xmark\ }  &  13.8    \\
\hline
\multicolumn{2}{l|}{\text {\  Wide-range Aug.\ \ }} &  {\ \ \cmark}  &  {\ \  \xmark\ }  &  14.6    \\
\multicolumn{2}{l|}{\text {\  Vanilla DA \ \ }}     &  {\ \ \cmark\ }  &  {\ \  \xmark\ }  &  15.1    \\
\multicolumn{2}{l|}{\text {\  Depth-Supv DA\ \ }}   &  {\ \ \cmark\ }  &  {\ \  \cmark\ }  &  16.5     \\
\multicolumn{2}{l|}{\text {\ Input-fusion Teacher \ \ }}  &  {\ \ \cmark}  &   {\ \  \cmark\ }  &  15.1   \\
\hline
\multicolumn{2}{l|}{\textbf {\  \ourwork (ours)\ \ }}                        &  {\ \  \cmark\ }  &  {\ \  \cmark\ }  & \textbf{19.2}\\
\hline
\end{tabu}} 
\vspace{-2mm}
\end{table}


\subsection{Lyft-to-nuScenes Adaptation } Due to page limit, Table~\ref{tab:n2l} in the main paper only shows the results under the nuScenes $\rightarrow$ Lyft setting. Here we also present the results under the Lyft $\rightarrow$ nuScenes adaptation setting for completeness. As shown in Table~\ref{tab:supp-l2n}, we can also observe that our \ourwork model achieves the best performance. Like previous scenarios, the baseline approaches, MonoLayout, OFT, and LSS without domain adaptation suffer from performance degradation due to domain shift. And compared with other baselines with domain adaptation or cross-modality learning, our \ourwork performs better in leveraging and adapting LiDAR information.

\begin{table*}[!t]
\caption{Different types of domain gaps react unevenly to different model designs. Direct depth supervision and the middle layer feature alignment block provide a larger improvement under the Day $\rightarrow$ Night setting than the Dry $\rightarrow$ Rain setting.} \label{tab:ablation_different_type}
\vspace{-1mm}
\centering
\resizebox{0.75\textwidth}{!}{
\begin{tabu}{@{}l|l|c|c@{}}
\hline
\multicolumn{2}{l|}{\textbf {\ \ \ourwork Designs\ \ }} & {\textbf {\ \ Day $\rightarrow$ Night\ \ }} & {\textbf {\ \ Dry $\rightarrow$ Rain\ \ }}  \\
\hline
\multicolumn{2}{l|}{\text {\ \ Image-only Baseline\ \ }}                         &  11.2  &  28.3     \\
\multicolumn{2}{l|}{\text {\ \ LiDAR Teacher Feature Supervision\ \ }}           &  14.9 \ \ \textit{(+3.7)}  & 29.5 \ \ \textit{\textbf{(+1.2)}}     \\
\multicolumn{2}{l|}{\text {\ \ LiDAR Teacher Feature + Depth Supervision\ \ }}   &  17.0 \ \ \textit{\textbf{(+5.8)}}  &  29.6 \ \ \textit{\textbf{(+1.3)}}   \\ 
\hline\hline
\multicolumn{2}{l|}{\text {\ \ Without Domain Alignment\ \ }}                   &  7.1  &  28.1   \\
\multicolumn{2}{l|}{\text {\ \ Feature Alignment at Final layer\ \ }}           &  12.2 \ \ \textit{(+5.1)}  &  29.6 \ \ \textit{\textbf{(+1.5)}}   \\
\multicolumn{2}{l|}{\text {\ \ Feature Alignment at Mid + Final layer\ \ }}     &  17.0 \ \ \textit{\textbf{(+9.9)}}  &  29.3 \ \  \textit{(+1.2)}\\ 
\hline
\end{tabu}} 
\vspace{-0mm}
\end{table*}


\begin{table*}[!t]
\hspace{0.03\linewidth}
\begin{minipage}{0.45\linewidth}
\centering
\vspace{2mm}
\caption{\ourwork achieves the best performance under simultaneous modality and domain shift for 3D object detection. Domain shift is \textbf{Singapore $\rightarrow$ Boston}.} \label{tab:detection-msds}
\vspace{-1mm}
\resizebox{\textwidth}{!}{
\begin{tabu}{@{}l|c|c@{}}
\hline
{\textbf { Modality-shift + Domain Shift }} & {mAP $\uparrow$} & {NDS $\uparrow$\ \ }\\
\hline
{\text {\ LSS~\cite{philion2020lift}}\ \ \ }  &  16.0  &  20.3 \\
{\text {\  MonoDistill~\cite{MonoDistill}\ \ }}  &  16.5  &  21.9 \\
{\text {\  Depth-Supv DA\ \ }}         &  19.1  &  23.5 \\
\hline
{\textbf {\  \ourwork\ \ }}         &  \textbf{22.5}  &  \textbf{26.1}  \\
\hline
\end{tabu}} 
\end{minipage}
\hspace{0.06\linewidth}
\begin{minipage}{0.4\linewidth}
\centering
\vspace{2mm}
\caption{\ourwork achieves great performance under only the modality shift for 3D object detection. EfficientNet and ResNet50 are backbones for image feature extraction.}
\label{tab:detection-ms}
\vspace{-1mm}
\resizebox{\textwidth}{!}{
\begin{tabu}{@{}l|c|c@{}}
\hline
{\textbf { Modality-shift Only }} & {\ mAP $\uparrow$\ } & {NDS $\uparrow$\ \ \ }\\
\hline
{\text {\  Set2Set~\cite{wang2021object}\ \ }} & {33.1}  &  {41.0} \\
{\text {\  MonoDistill~\cite{MonoDistill}\ \ }} & {34.3} &  {41.2} \\
\hline
{\text {\  \ourwork-EfficientNet\ \ }}  &  {34.5}  &  {41.5} \\
{\textbf {\  \ourwork-ResNet50\ \ }}  &  \textbf{35.2}  &  \textbf{42.4} \\
\hline
\end{tabu}} 
\end{minipage}
\vspace{-2mm}
\end{table*}


\subsection{Effect of Designs on Different Domain Gaps} 

We find in our experiments that different types of domain gaps react unevenly to different model designs. As shown in Table~\ref{tab:ablation_different_type}, for the Day $\rightarrow$ Night setting which has less domain shift in the LiDAR point cloud modality than the camera modality, the depth supervision and the middle layer feature alignment block provide a large improvement on top of other modules. By contrast, under the Dry $\rightarrow$ Rain setting, where domain gap is larger for point clouds than images, no significant improvement is achieved by using these two components. This further validates the necessity of using multiple modalities under domain adaptation settings, which can effectively improve the algorithm robustness under different domain shifts.

\subsection{3D Detection Results and Comparisons}\label{sec:exp:detection}

\noindent \textbf{Baselines. } In addition to the BEV segmentation task, we compare \ourwork with state-of-the-art cross-modality 3D object detection models, MonoDistill~\cite{MonoDistill} and Set2Set~\cite{wang2021object}. The original MonoDistill was designed for the single-camera setting on the KITTI dataset, so we extend it into a multi-camera setting for a fair comparison. We evaluate with mean Average Precision (mAP) and Nuscenes Detection Score (NDS) metrics~\cite{Caesar2019nuScenes}.

\vspace{2mm}
\noindent \textbf{Modality-Shift + Domain-Shift. } As shown in Table~\ref{tab:detection-msds}, under concurrent modality and domain gaps, \ourwork outperforms all previous baselines by a large margin, showing that it is a more robust and trustworthy model in real-world scenarios.

\vspace{2mm}
\noindent \textbf{Modality-Shift Only. } As shown in Table~\ref{tab:detection-ms} on the right, our \ourwork model achieves the best performance. By using ResNet-50, the same image feature extractor backbone as MonoDistill, we can achieve more than 1\% improvement in mAP and NDS metrics. Moreover, our model still runs 42ms per frame during inference, faster than MonoDistill which runs 80ms per frame.

\subsection{Solving Scale Problem} 

Scale ambiguity is an inherent problem for monocular depth estimation. We solve this problem by using the random cropping and scaling augmentation strategy during training, and also by matching the FoV (Field-of-View) of two domains using their intrinsic matrices. The augmentation increases the robustness of the depth prediction model in scale difference. And the FoV matching scales the images in the target domain to match the FoV of the source domain. This ensures that one object looks approximately the same size across images, if it is of the same distance to the ego vehicle in source and target domains, thus reducing the scale ambiguity. In Table~\ref{tab:supp-scale}, we provide an ablation study of scale augmentation and FoV matching in {nuScenes-to-Lyft} adaptation. As we can see, they both improve our final model performance.

\begin{table}[!t]
\caption{\ourwork achieves great performance with scaling augmentation and FoV matching for \textit{{nuScenes-to-Lyft}} domain gaps in vehicle IoU.} \label{tab:supp-scale}
\vspace{0mm}
\centering
\resizebox{0.46\textwidth}{!}{
\begin{tabu}{@{}l|c|c||c@{}}
\hline
{\textbf { nuScenes $\rightarrow$ Lyft }} & {\ {Scaling Aug.}\ } & {\ {Match FoV}\ } & {\ Vehicle IoU\ } \\
\hline
{\text {\ None}\ \ \ } &    &    &  23.5    \\
{\text {\ with scaling aug.}\ \ \ }   &  {\ \ \checkmark\ }  &    &  23.8    \\ 
{\text {\ with both}\ \ \ }           &  {\ \ \checkmark\ }  &  {\ \  \checkmark\ }  &  24.4    \\
\hline
\end{tabu}} 
\vspace{-2mm}
\end{table}


From Table~\ref{tab:supp-scale}, we also notice that even without the two methods mentioned above, the model still performs fairly well, compared with the baseline methods in Table~\ref{tab:n2l}. This is because in driving scenarios, the camera FoV, the context in the images, and the depth distribution of the images have a relatively strong prior – they do not have a strong discrepancy even across different driving scenarios (domains). This is different from more general depth estimation scenarios, where objects can have drastically different depth distribution and the intrinsic matrix can have big differences from image to image.

\begin{table*}[!t]
\caption{Our proposed modules achieve great performance in both depth estimation and semantic segmentation tasks.} \label{tab:supp-separate}
\vspace{0mm}
\centering
\resizebox{0.9\textwidth}{!}{
\begin{tabu}{@{}l|c|c@{}}
\hline
{\textbf { Day $\rightarrow$ Night }} & {\ {Depth Estimation (CEE)}\ } & {\ Semantic Segmentation (IoU)\ } \\
\hline
{Direct Inference\ \ \ } &  2.56  &  27.5    \\
{Adversarial Learning (AL)\ \ \ }  &  1.97  &  31.8    \\ 
{Lidar-Teacher Supervision + AL\ \ \ }           &  \textbf{1.41}  &  \textbf{32.1}    \\
\hline
\end{tabu}} 
\vspace{-2mm}
\end{table*}

\begin{table*}[!t]
\caption{Naive LiDAR supervision cannot help the final perception under cross-domain and cross-modality scenarios.} \label{tab:supp-lidar-supp-other}
\vspace{0mm}
\centering
\resizebox{0.9\textwidth}{!}{
\begin{tabu}{@{}l|c|c||c@{}}
\hline
{\textbf{ Method }} & {\textbf{Boston $\rightarrow$ Singapore}} & \textbf{Dry $\rightarrow$ Rain} & \textbf{nuScenes$\rightarrow$Lyft} \\
\hline
{\text {\ LSS~\cite{philion2020lift} Baseline}\ \ \ } &  17.5  &  27.9  &  20.6    \\
{\text {\ Baseline + naïve LiDAR supervision}\ \ \ }   &  17.5 (-0.0)  &  28.2 (+0.3)  &  20.3 (-0.2)    \\ 
{\text {\ \ourwork (Full Model)}\ \ \ }           & \textbf{20.5 (+3.0)}  &  \textbf{29.6 (+1.7)}  &  \textbf{24.4 (+3.9)}    \\
\hline
\end{tabu}} 
\vspace{-2mm}
\end{table*}

\subsection{Ablation in Semantic Segmentation and Depth Prediction} 
From the task perspective, there are two domain gaps in this task, one in the semantic segmentation task and the other in the depth prediction task. In Table~\ref{tab:supp-separate}, we show the result of depth estimation and semantic segmentation alone in the Day-to-Night scenario.

For depth estimation, we report the cross-entropy error (CEE) with the ground truth we described in Sec~\ref{sec:method_teacher}, because each pixel will have multiple depth values. For this task, the LiDAR teacher supervision refers to $L_\mathrm{dp}$ in our main pipeline. We observe that our method significantly improves the depth estimation metrics by 44.9\%.

For semantic segmentation, we report the IoU for the vehicle class. To remove the effect of depth estimation, we use a pre-trained depth estimation model and fix its weight. For this task, the LiDAR teacher supervision refers to $L_\mathrm{T}$ in our main pipeline. We observe that the \ourwork architecture also improves the baseline method by 16.7\%.

\subsection{Results of Naive LiDAR Supervision Under Other Adaptation Settings} 
In Table~\ref{tab:supp-lidar-supp-other}, we present the ablation study of direct LiDAR supervision under different adaptation scenarios. We observe that in all scenarios, naive LiDAR supervision cannot lead to better performance against the baseline.

When the source and target domains have larger visual gaps (Day$\rightarrow$Night, nuScenes$\rightarrow$Lyft), the naive supervision leads to worse results, and when the gap is smaller (Boston→Singapore, Dry→Rain), the LiDAR supervision results in on-par or only slightly better performance. The reason is that although the LiDAR sensor in the source domain provides 3D knowledge to the model, it also increases the domain discrepancy between the source and the target (the model has to adapt to the additional modality shift), which hurts the model performance instead.

\section{Visualization on Failure Cases}

\begin{figure*}[!t]
    \centering
    \includegraphics[trim=180 70 180 50,clip, width=0.85\linewidth]{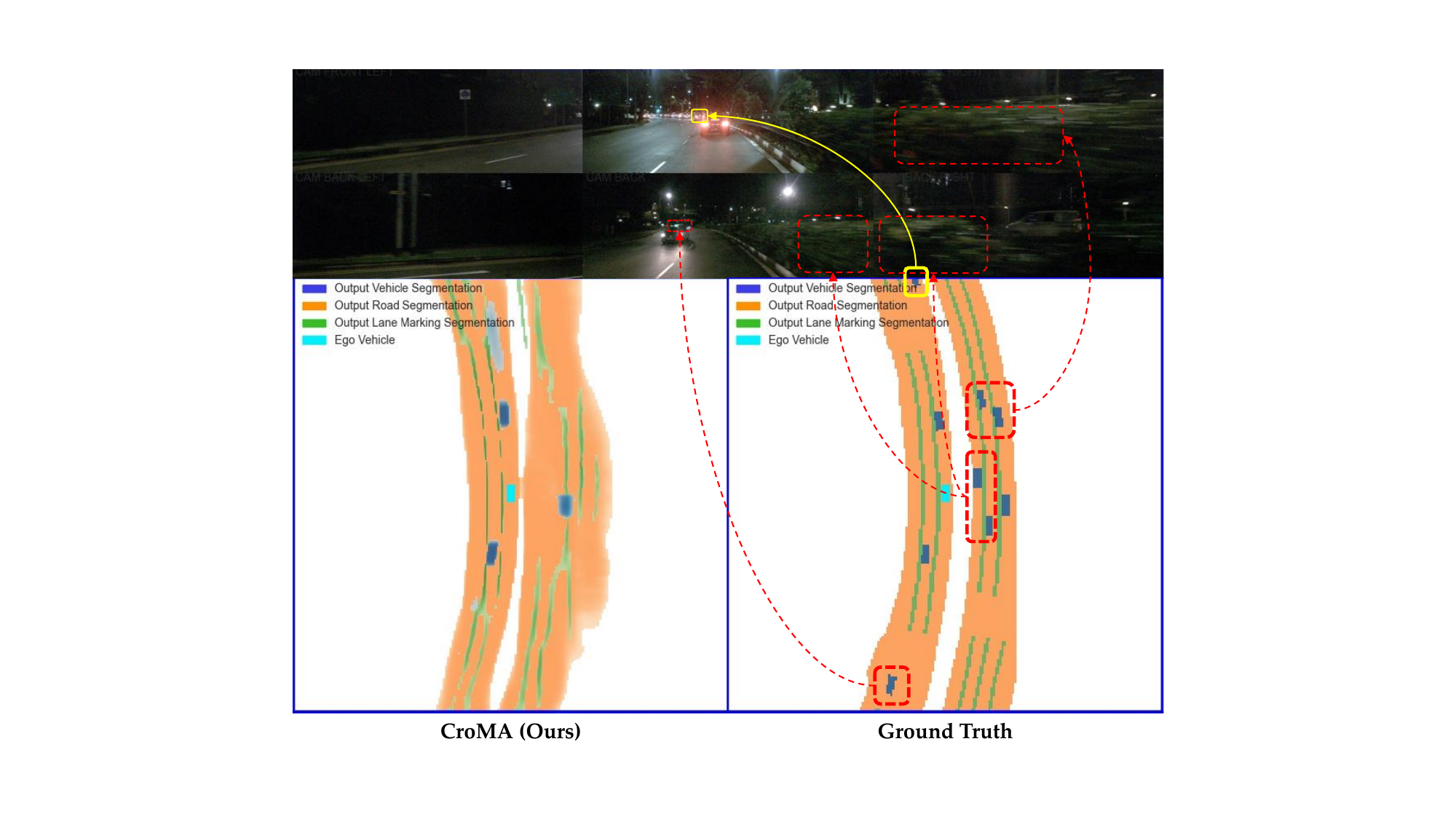}
    
    \vspace{3mm}
    \includegraphics[trim=180 70 180 50,clip, width=0.85\linewidth]{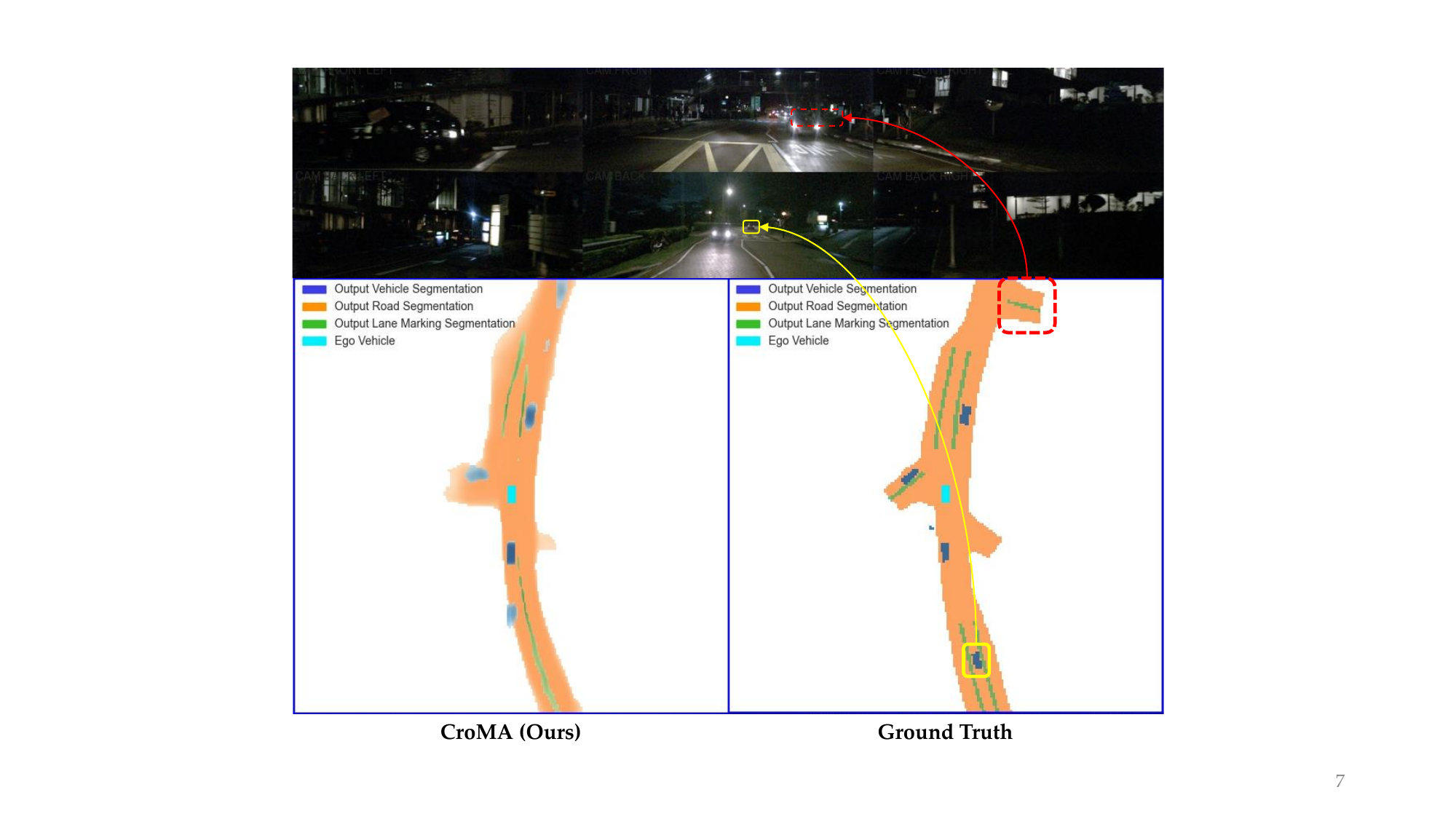}
    \vspace{-1mm}
    \caption{\textbf{Visualization of failure cases of \ourwork} (model is trained with daytime data, and validated with night data). \textit{Up}: Input images. \textit{Left}: Prediction from \ourwork. \textit{Right}: Ground truth. We notice that major failure cases of \ourwork are \textcolor{myyellow}{far distance} and \textcolor{red}{occlusions} of objects and regions.}
    \label{fig:supp_failure}
\end{figure*}

As shown in Figure~\ref{fig:supp_failure}, we visualize the failure cases of our \ourwork model. We notice that most failure cases come from far distance, or heavy occlusion (which are typical failure cases for baselines as well). Faraway objects are known to be typically hard cases for monocular 3D perception~\cite{xu2018multi,wang2019pseudo,brazil2019m3d,wang2021fcos3d,wang2022probabilistic}, because the ambiguity of object depth from images becomes larger as the distance increases. This can be potentially alleviated by using a smaller downsampling rate when extracting the image features. As for objects inside the red dashed boxes in Figure~\ref{fig:supp_failure}, they can be seen in the LiDAR sensor due to the higher deployment position. But in cameras, the objects are almost invisible due to the occlusion of vegetation, structures, or other vehicles. The occlusion problem can be potentially addressed if we have access to additional sensors during evaluation. As future work, we will also try to alleviate the occlusion problem in monocular settings by leveraging temporal information.

\end{document}